\documentclass[final,3p,times,twocolumn]{elsarticle}
\usepackage{graphicx}
\usepackage{amssymb}
\usepackage{amsmath}
\usepackage{amsthm}
\usepackage{multirow} 
\usepackage{xspace}

\usepackage{lineno}

\begin{document}
 
 \begin{frontmatter}
  
  
  
  
  \title{P3DC-Shot: Prior-Driven Discrete Data Calibration for \\
  Nearest-Neighbor Few-Shot Classification}
  
  
  \author[1]{Shuangmei Wang \corref{cor1}}
  
  \affiliation[1]{organization={Jilin University},
   addressline={No. 2699 Qianjin Street}, 
   city={Changchun},
   postcode={130012}, 
   country={China}}
   
  \affiliation[2]{organization={Engineering Research Center of Knowledge-Driven Human-Machine Intelligence, MOE},
    addressline={No. 2699 Qianjin Street}, 
   city={Changchun},
    postcode={130012}, 
   country={China}}

  \author[1,2]{Rui Ma \corref{cor1}}
  
  \author[1,2]{Tieru Wu \corref{cor2}}

  \author[1]{Yang Cao \corref{cor2}}

            \cortext[cor1]{Co-first authors.}
            \cortext[cor2]{Corresponding authors.}

  \nonumnote{
   $\bullet$ This work is supported in part by the National Key Research and Development
   Program of China (Grant No. 2020YFA0714103) and the National Natural Science
   Foundation of China (Grant No. 61872162 and 62202199).
  }

   \begin{abstract}  
   Nearest-Neighbor (NN) classification has been proven as a simple and effective approach for few-shot learning.
   The query data can be classified efficiently by finding the nearest support class based on features extracted by pretrained deep models.
   However, NN-based methods are sensitive to the \textit{data distribution} and may produce false prediction if the samples in the support set happen to lie around the distribution boundary of different classes.
   To solve this issue, we present P3DC-Shot, an improved nearest-neighbor based few-shot classification method empowered by prior-driven data calibration.
   Inspired by the distribution calibration technique which utilizes the distribution or statistics of the base classes to calibrate the data for few-shot tasks, we propose a novel \textit{discrete} data calibration operation which is more suitable for NN-based few-shot classification.
   Specifically, we treat the prototypes representing each base class as priors and calibrate each support data based on its similarity to different base prototypes.
   Then, we perform NN classification using these discretely calibrated support data.
   Results from extensive experiments on various datasets show our efficient non-learning based method can outperform or at least comparable to SOTA methods which need additional learning steps.
  \end{abstract}
  
 \end{frontmatter}

 \section{Introduction}
 Deep learning has triggered significant breakthroughs in many computer vision tasks, such as image classification \cite{simonyan2014very, 2015ImageNet, 2016deep}, object detection \cite{2015Fast,2017Faster,redmon2016you}, and semantic segmentation \cite{2015Fully, 2017Mask, 2018DeepLab} etc. 
 One key factor for the success of deep learning is the emergence of large-scale datasets, e.g., ImageNet \cite{2015ImageNet}, MSCOCO \cite{lin2014microsoft}, Cityscapes \cite{cordts2016cityscapes}, just to name a few. 
 However, it is difficult and expensive to collect and annotate sufficient data samples to train a deep model with numerous weights.
 The data limitation has become a main bottleneck for more broader application of deep leaning, especially for the tasks involving rarely seen samples. 
 On the other hand, human can learn to recognize novel visual concepts from only a few samples.
 There is still a notable gap between human intelligence and the deep learning based artificial intelligence.
 Few-shot learning (FSL) aims to learn neural models for novel classes with only a few samples. 
 Due to its ability for generalization, FSL has attracted extensive interests in recent years \cite{wang2020generalizing,2020Learning,huang2022survey}. 
 
 Few-shot classification is the most widely studied FSL task which attempts to recognize new classes or classify data in an unseen \textit{query set}.
 Usually, few-shot classification is formulated in a meta-learning framework \cite{matching, prototypical, relation,optimization,MAML,Jamal_2019_CVPR,LEO,DCO,meta-baseline}.
 In the meta-training stage, the N-way K-shot episodic training paradigm is often employed to learn generalizable classifiers or feature extractors for data of the \textit{base classes}. 
 Then, in the meta-testing stage, the meta-learned classifiers can quickly adapt to a few annotated but unseen data in a \textit{support set} and attain the ability to classify the novel query data.
 Although meta-learning has shown the effectiveness for few-shot classification, it is unclear how to set the optimal class number (N) and per-class sample number (K) when learning the classifiers.
 Also, the learned classifier may not perform well when the sample number K used in meta-testing does not match the one used in the meta-training \cite{theoretical}.
 
\begin{figure}[t]
  \centering
  \includegraphics[width=0.95\linewidth]{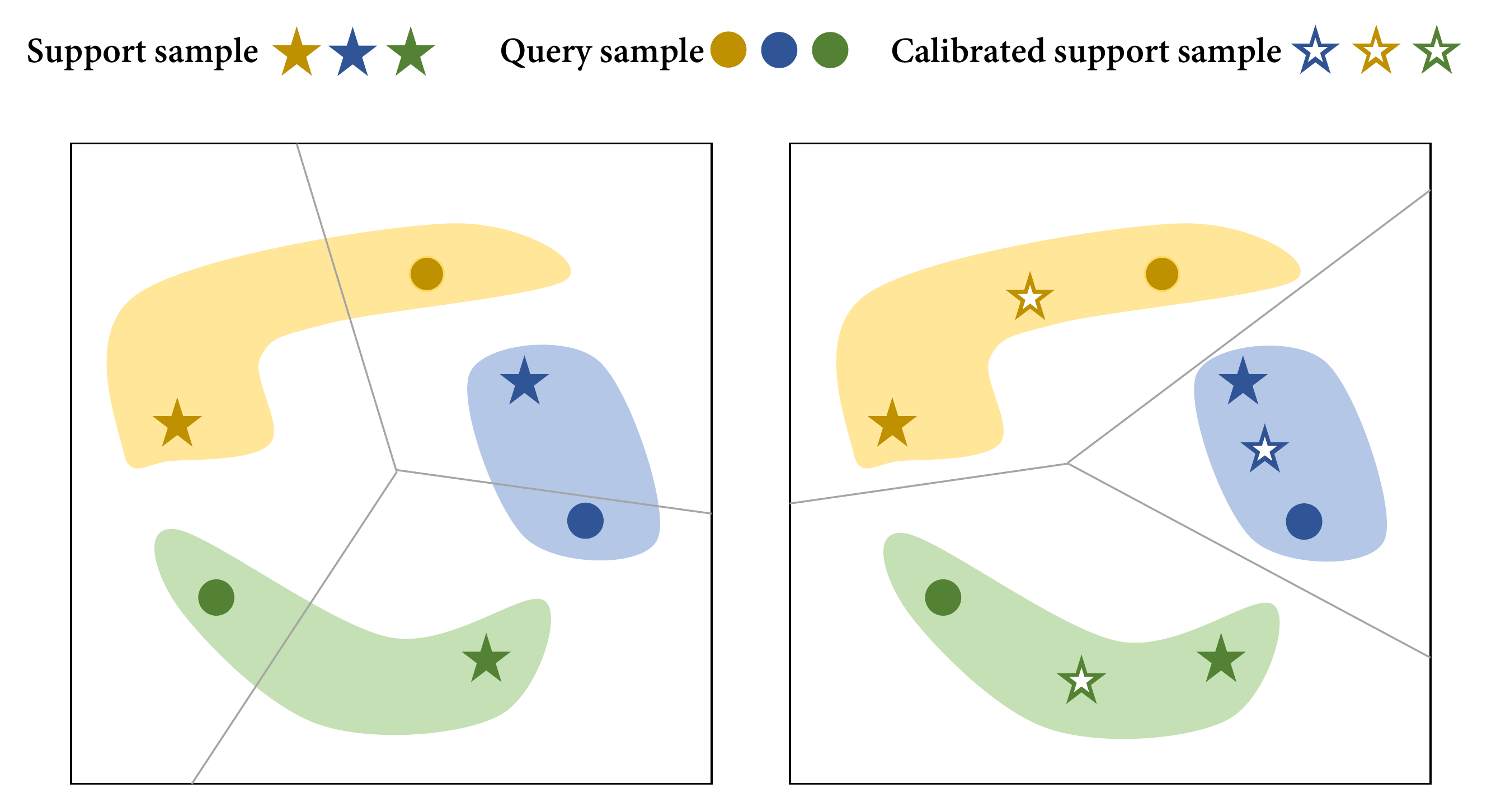}
  \caption{When samples in the support set lie around the distribution boundary of different classes, the NN classifier may produce false prediction.
   By performing discrete calibration for each support sample using priors from the base classes, the calibrated support data is transformed closer to the actual class centroid and can lead to less-biased NN classification.
   The colored regions represent the underlying data distribution of different classes.
   The gray lines are the predicted decision boundaries by the NN classifier.
  }
  \label{calibrated}
 \end{figure}

 On the other hand, nearest-neighbor (NN) based classification has been proven as a simple and effective approach for FSL.
 Based on features obtained from the meta-learned feature extractor \cite{matching,prototypical} or the pretrained deep image models \cite{simpleshot},
 the query data can be efficiently classified by finding the nearest support class.
 Specifically, the prediction is determined by measuring the similarity or distance between the query feature and the prototypes (i.e., average or centroid) of the support features.
 From the geometric view, NN-based classification can be solved using a Voronoi Diagram (VD) which is a partition of the space formed by the support features \cite{aurenhammer1991voronoi,chen2017clustering}.
 Given a query feature, its class can be predicted by computing the closest Voronoi cell that corresponds to a certain support class.
 With proper VD construction and feature distance metrics, the state-of-the-art performance can be achieved for few-shot classification \cite{ma2022few}.
 However, due to the limited number of support samples, NN-based few-shot classification is sensitive to the distribution of the sampled data and may produce false prediction if the samples in the support set happen to lie around the distribution boundary of different classes (see Figure \ref{calibrated} left).
 
 To solve above issues, various efforts have been paid to more effectively utilize the knowledge or priors from the base classes for few-shot classification.
 One natural way is to learn pretrained classifiers or image encoders with the abundant labeled samples of base classes and then adapt them the novel classes via transfer learning \cite{s2m2,rethinking,closer,meta-baseline}.
 Meanwhile, it has been shown that variations in selecting the base classes can lead to different performance on the novel classes \cite{Ge_2017_CVPR,OthmanSbai2020ImpactOB, select} and how to select the base classes for better feature representation learning still needs more investigation.
 On the other hand, a series of works \cite{one-shot,liu2020prototype,guo2022learning,DC} perform data calibration to the novel classes so that the results are less affected by the limited number of support samples.
 One representative is Distribution Calibration (DC) \cite{DC} which assumes the features of the data follow the Gaussian distribution and transfers the statistics from the similar base classes to the novel classes.
 Then, DC trains a simple logistic regression classifier to classify the query features using features sampled from the calibrated distributions of the novel classes.
 Although DC has achieved superior performance than previous meta-learning \cite{MAML,LEO,DCO} or transfer-learning \cite{s2m2,rethinking,closer,meta-baseline} based methods, it relies on the strong assumption for Gaussian-like data distribution and it cannot be directly used for NN-based few-shot classification.

 In this paper, we propose P3DC-Shot, an improved NN-based few-shot classification method that employs prior information from base classes to \textit{discretely} calibrate or adjust the support samples so that the calibrated data is more representative for the underlying data distribution (Figure \ref{calibrated} right).
 Our main insight is even the novel classes have not been seen before, they still share similar features to some base classes, and the prior information from the base classes can serve as the \textit{context} data for the novel classes.
 When only a few support samples are available for the novel classes, performing prior-driven calibration can alleviate the possible bias introduced by the few-shot support samples.
 With the calibrated support samples, the query data can be more accurately classified by a NN-based classifier.
 
 Specifically, for the prior information, we compute the prototype, i.e., the average of features, for each base class.
 Then, we propose three different schemes for selecting the similar prototypes to calibrate the support data.
 Firstly, we propose the sample-level calibration which selects the top $M$ most similar base prototypes for each support sample and then apply weighted averaging between each support sample and selected prototypes to obtain the calibrated support sample.
 Secondly, to utilize more context from the base classes, we propose the task-level calibration which combines the most similar base prototypes for each support sample into a union and performs the calibration for the support samples using each prototype in the union.
 In addition, we propose a unified calibration scheme that combines the two above schemes so that the calibration can exploit different levels of prior information from the base classes.
 To utilize the calibrated support samples for the NN-based classification, we further obtain the prototypes of the support class using an attention-weighted averaging, while the attention weights are computed between the query sample and each calibrated support sample.
 Finally, the classification of a query sample is simply determined by finding its nearest support prototype measured by the cosine similarity. 
 
 Comparing to DC, our P3DC-Shot adopts the similar idea of transferring the information or statistics from the base classes to the novel classes.
 The key difference is our data calibration is performed on each individual support sample rather than the distribution parameters and we employ the NN-based classification instead of the learned classifier as in DC.
 Comparing to other NN-based few-shot classification methods such as SimpleShot \cite{simpleshot}, since our support data is calibrated, the NN classification is less affected by the sampling bias for the support data, e.g, the calibrated data is more likely to be close to the center of the corresponding novel class.
 We conduct extensive comparisons with recent state-of-the-art few-shot classificaiton methods on miniImageNet \cite{2015ImageNet}, tiredImageNet \cite{M2018Meta} and CUB \cite{CUB} and the results demonstrate the superiority and generalizability of our P3DC-Shot.
 Ablation studies on different calibration schemes, i.e., different weights between the sample-level and task-level calibration also show the necessity of combining two schemes for better results.
 
 In summary, our contributions are as follows:

 \begin{enumerate}
  \item We propose P3DC-Shot, a prior-driven discrete data calibration strategy for nearest-neighbor based few-shot classification to enhance the model's robustness to the distribution of the support samples.
  \item Without additional training and expensive computation, the proposed method can efficiently calibrate each support sample using information from the prototypes of the similar base classes.
  
  \item We conduct extensive evaluations on three discrete calibration schemes on various datasets and the results show our efficient non-learning based method can outperform or at least comparable to SOTA few-shot classification methods.
 \end{enumerate}

 \section{Related Work}
 In this section, we first review the representative meta-learning and transfer learning based few-shot classification techniques.
 Then, we summarize the nearest-neighbor and data calibration based approaches which are most relevant to our P3DC-Shot.
 
 \textbf{Meta-learning based few-shot classification.}
 Meta-learning \cite{hospedales2021meta} has been widely adopted for few-shot classification.
 The core idea is to leverage the episodic training paradigm to learn generalizable classifiers or feature extractors using the data from the base classes in an optimization-based framework \cite{optimization,MAML,Jamal_2019_CVPR,LEO,DCO}, as well as learn a distance function to measure the similarity between the support and query samples through metric-learning \cite{koch2015siamese,matching,relation,xu2021attentional,liu2022dmn4,guo2022learning}.
 For example, MAML \cite{MAML} is one of the most representative optimization-based meta-learning method for few-shot classification and its goal is to learn good network initialization parameters so that the model can quickly adapt to new tasks with only a small amount of new training data from the novel classes.
 For metric-learning based methods such as the Matching Networks \cite{matching}, Prototypical Networks \cite{prototypical} and Relation Networks \cite{relation}, the network is trained to either learn an embedding function with a given distance function or learn both the embedding and the distance function in a meta-learning architecture.
 Unlike the optimization and metric-learning based methods which require sophisticated meta-learning steps, our method can directly utilize the features extracted by the pretrained models and perform the prior-driven calibration to obtain less-biased support features for classification.
 
 \textbf{Transfer learning based few-shot classification.}
 Transfer learning \cite{torrey2010transfer,tan2018survey,zhuang2020comprehensive} is a classic machine learning or deep learning technique that aims to improve the the learning of a new task through the transfer of knowledge from one or more related tasks that have already been learned.
 Pretraining a deep network on the base dataset and transferring knowledge to the novel classes via fine-tuning \cite{closer,Dhillon2020A,rethinking} has been shown as the strong baseline for the few-shot classification.
 To learn better feature representations which can lead to improved few-shot fine-tuning performance, Mangla et al. \cite{s2m2} propose S2M2, the Self-Supervised Manifold Mixup, to apply regularization over the feature manifold enriched via the self-supervised tasks.
 In addition to training new linear classifiers based on the pretrained weights learned from the base classes, Meta-Baseline \cite{meta-baseline} performs meta-learning to further optimize the pretrained weights for few-shot classification.
 On the other hand, it has been shown the results of the transfer learning based methods depend on different selections of the base classes for pretraining \cite{Ge_2017_CVPR,OthmanSbai2020ImpactOB}, while how to select the base classes to achieve better performance is still challenging \cite{select}.
 In comparison, our P3DC-shot does not need the additional cost for feature representation learning and can more effectively utilize the base classes in a NN-based classification framework.
 
 \textbf{Nearest neighbor based few-shot classification.}
 NN-based classification has also been investigated for few-shot classification.
 The main idea is to compute the prototypes of the support samples, i.e., the mean or centroid of the support features, and classify the query sample using metrics such as L2 distance, cosine similarity or a learned distance function.
 In SimpleShot \cite{simpleshot}, it shows nearest neighbor classification with features simply normalized by L2 norm and measured by Euclidean distance can achieve competitive few-shot classification results. 
 Instead of performing nearest neighbor classification on the image-level features, Li et al. \cite{li2019dn4} introduces a Deep Nearest Neighbor Neural Network which performs nearest neighbor search over the deep local descriptors and defines an image-to-class measure for few-shot classification.   
 From a geometric view, Ma et al. \cite{CIVD} utilize the Cluster-induced Voronoi Diagram (CIVD) to incorporate cluster-to-point and cluster-to-cluster relationships to the nearest neighbor based classification.
 Similar to above methods, our method is based on the nearest prototype classification, while we perform the prior-driven data calibration to obtain less-biased support data for the prototype computation.
 Meanwhile, computing the attentive or reweighted prototypes \cite{wu2020,ji2021,wang2022} that are guided by the base classes or query samples has also been investigated recently.
 We follow the similar idea and compute the attention-weighted prototypes for NN-based classification.

 \textbf{Data calibration for few-shot classification.}
 Due to the limited number of samples, the prototypes or centroids computed from the few-shot support data may be biased and cannot represent the underlying data distribution.
 Simply performing NN-based classification on these biased prototypes will lead to inaccurate classification.
 Several methods have been proposed to calibrate or rectify the data to obtain better samples or prototypes of the support class \cite{one-shot,liu2020prototype,guo2022learning,xu2022alleviating,DC}.
 Using the images in the base classes, RestoreNet \cite{one-shot} learns a class agnostic transformation on the feature of each image to move it closer to the class center in the feature space.
 To reduce the bias caused by the scarcity of the support data, Liu et al., \cite{liu2020prototype} employ the pseudo-labeling to add unlabelled samples with high prediction confidence into the support set for prototype rectification.
 In \cite{guo2022learning}, Guo et al. propose a Pair-wise Similarity Module to generate calibrated class centers that are adapted to the query sample.
 Instead of calibrating individual support samples, Distribution Calibration (DC) \cite{DC} aims to calibrate the underlying distribution of the support classes by transferring the Gaussian statistics from the base classes.
 With sufficient new support data sampled from the calibrated distribution, an additional classifier is trained in \cite{DC} to classify the query sample.
 In contrast to these methods, we do not require additional training or assumption of the underlying distribution.
 Instead, we directly use the prototypes of the base classes to calibrate each support sample individually and we adopt the NN-based classification which makes the whole pipeline discrete and efficient.
 One recent work that is similar to ours is Xu et al. \cite{xu2022alleviating} which proposes the Task Centroid Projection Removing (TCPR) module and transforms all support and query features in a given task to alleviate the sample selection bias problem.
 Comparing to \cite{xu2022alleviating}, we only calibrate the support samples using the priors from the base classes and keep the query samples unchanged.
 
 \section{Method}
 
 \begin{figure*}[t]
  \centering
  \includegraphics[width=0.95\linewidth]{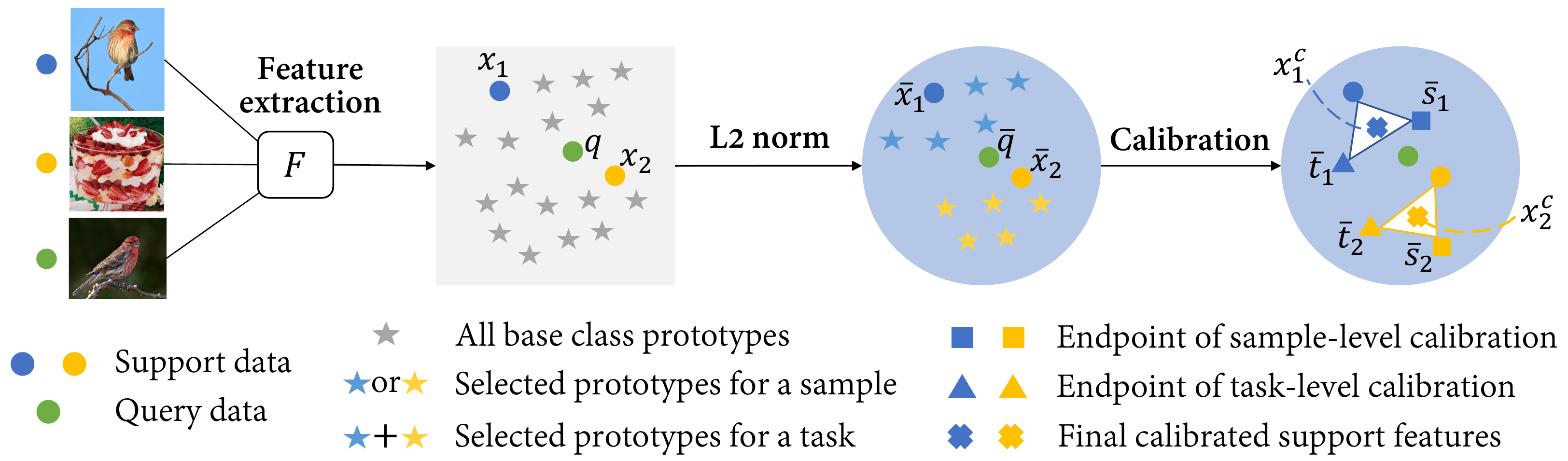}
  \caption{An illustration of the P3DC-Shot pipeline for the 2-way 1-shot scenario.
   Note that the direct interpolation of the three triangle vertices return a feature on the triangle plane.
   After normalization, the final calibrated features $\bar{x}_1^u$ and $\bar{x}_2^u$ are on the hypersphere of the normalized space.
  }
  \label{framework}
 \end{figure*}
 
 To effectively utilize the prior knowledge from the base classes, we first propose two independent calibration strategies, i.e., sample-level calibration and task-level calibration, which exploit different levels of information from the base classes.
 Then, we combine the sample-level and task-level calibration together to obtain the final calibrated support samples which will be used for the nearest neighbor classification.
 
 Figure \ref{framework} shows an illustration of the P3DC-Shot pipeline.
 Given a pretrained feature extractor $F$ and a set of prototypes of base classes, we perform the prior-driven discrete calibration to the normalized features of the support data.
 Initially, the query sample in green is closer to the support sample in yellow.
 After the proposed calibration using the related base class prototypes, the query sample becomes closer to the calibrated support sample in blue.
 In the following, we provide technical details of the P3DC-Shot for few-shot classification.  
 
 \subsection{Problem Statement}
 In this paper, we focus on the few-shot image classification which aims to classify the new image samples from the novel classes with just a few labeled image samples.
 Normally, the new data sample is called a query sample and the labelled samples are called support samples.  
 With the aid of a set of base classes represented by their prototypes $P^{b}= \{ p_i^b  \}_{i=1}^{n_b}$, our goal is to calibrate the support samples from novel-class so that they can be better matched with the query samples by a nearest neighbor classifier.
 Here, all data samples are represented by the features computed from a pretrained feature extractor $F(\cdot): X \rightarrow R^d$, while $X$ is the domain of the image space and $d$ is the dimension of the feature space;
 $p_i^b$ is the prototype of a base class, which is computed as the average feature of the samples within the class;
 $n_b$ is the number of all base classes.
 For simplicity, we directly use $x_i$ to represent the feature $F(x_i)$ of an image $x_i$.
 
 We follow the conventional few-shot learning setting, i.e., build a series of N-way K-shot tasks where $N$ is the number of novel classes and $K$ is the number of support samples in each task.
 Formally, each task consists of a support set $\mathbf{S} = \{ (x_i, y_i)\}_{i=1}^{N \times K}$ and a query set $\mathbf{Q} = \{q_i\}_{i=N\times K +1}^{N \times K+ N \times Q}$.
 Here, $y_i$ is the label of the corresponding sample, which is known for the support set and unknown for the query set; $Q$ is the number of query sample for each novel class in the current task.
 Given a support feature $x_i$, we perform our prior-driven calibration to obtain the calibrated support feature $x_i^c=\mathcal{C}(x_i)$, where $\mathcal{C}(\cdot):R^d \rightarrow R^d$ conducts feature transformation based on the information from the base classes.
 Then, we predict the label of a query feature by performing nearest neighbor classification w.r.t the novel class prototypes computed from the calibrated support feature(s).
 
 \subsection{Prior-Driven Discrete Data Calibration}
 
 Before we perform calibration to the support data, we first apply L2 normalization to the support and query features.
 It is shown in SimpleShot \cite{simpleshot} that using L2-normalized feature with a NN-based classifier can lead to competitive results for few-shot classification. 
 Hence, we obtain $\bar{x}_i$ for a support feature $x_i$ by:
 \begin{equation}
  \bar{x}_i = normalize(x_i)= \frac{x_i}{\left\|x_i \right\|_2}.
 \end{equation}
 Similarly, the normalization of the query features are also computed: $\bar{q}_i=normalize(q_i)$.
 By working with the normalized features, we can obviate the absolute scales of the features and focus on the similarities and differences on their directions.
 Note that, the normalized features are used in the feature combination step (Eq. \ref{eq:sample-calib}, \ref{eq:task-calib} and \ref{eq:final-calib}) for obtaining the interpolation between the normalized features and in the NN-based classification step (Eq. \ref{eq:nn}) for performance improvement.

 Next, we propose the sample-level and task-level calibration, and their combination to utilize the priors from the base classes for obtaining the less-biased support features.  
 
 \subsubsection{Sample-Level Calibration }\label{Sample}
 According to previous works \cite{SR, DC} which also use the information from base classes for classifying the new classes, the base classes with higher similarities to the query classes are more important than other base classes.
 Hence, we first propose to perform calibration based on the top similar base classes for each support sample.
 Moreover, following DC \cite{DC}, we apply the Tukeys's Ladder of Powers transformation \cite{tukey} to the features of the support samples before the calibration:
 \begin{equation}
  \label{eq:transformation}
  \Tilde{x}_i =\left\{\begin{array}{ll}
   x^{\lambda}_i & \text { if } \lambda \neq 0 \\
   \log (x_i) & \text { if } \lambda=0 \\
  \end{array}\right.
 \end{equation}
 Here, $\lambda$ is a hyperparameter which controls the distribution of the transformed feature, with a smaller $\lambda$ can lead to a less skewed feature distribution.
 We set $\lambda = 0.5$ and obtain the transformed support feature $\tilde{x}_i$ from the original feature $x_i$.
 
 Then, we select the top $M$ base classes with higher similarities to a transformed support feature $\tilde{x}_i$:
 \begin{align}
  \label{eq:k-nearest}
  \Lambda_{i}^M &= \{ p_j^b|  j \in topM(\mathbb{S}_i) \}, \\
  \label{eq:distance}
  where \enspace \mathbb{S}_i &= \{<\Tilde{x}_i, p_j^b > | j\in \{1, \dots n_b\} \}. 
 \end{align}
 Here, $\Lambda_{i}^M$ stores the $M$ nearest base prototypes with respect to a transformed support feature vector $\Tilde{x}_i$;
 $topM(\cdot)$ is an operator that returns the index of top $M$ elements from $\mathbb{S}_i$, the similarity set of $\tilde{x}_i$, while the similarity between $\tilde{x}_i$ and a base prototype $p^b_j$ is computed by the inner product $<\cdot, \cdot>$. 
 In DC \cite{DC}, the distributions of the base and novel classes are assumed as Gaussian distribution and the statistics (mean and co-variance) of the base classes are used to calibrate the distribution of the novel classes.
 In contrast, we directly use the similar base prototypes to calibrate each support feature.
 Specifically, the calibration for $\tilde{x}_i$ driven by base prototypes $p_j^b \in \Lambda_{i}^M$ is computed as:
 \begin{equation}
  \label{eq:sample-pb}
  s_i = \tilde{x}_i +\sum_{j \in \Lambda_{i}^M } w_{ij} p_j^b,
 \end{equation}
 where the weights of the M nearest base classes prototypes in $\Lambda_{i}^M$ are obtained by applying Softmax to the similarities between $\tilde{x}_i$ and these prototypes:
 \begin{equation}
  \label{eq:weight}
  w_{ij} = \frac{e^{<\Tilde{x}_i, p_j^b >}}{\sum_{k \in \Lambda_{i}^M } e^{<\Tilde{x}_i, p_k^b >}}, j\in  \Lambda_{i}^M.
 \end{equation}
 It should be noted that, in Eq. \ref{eq:sample-pb}, the support feature $\tilde{x}_i$ is a transformed feature, while the base prototypes are in the original feature space.
 This setting is the same as DC does for calibrating the distribution of the novel classes and it can be understood as follows: 
 1) the transformation can initially reduce the skewness of the few-shot-sampled support features; 
 2) the term $w_{ij} p_j^b$ can be regarded as the projection of $\tilde{x}_i$ w.r.t prototype $p_j^b$;
 3) $\tilde{x}_i$ is calibrated based on its projects to all of its similar base prototypes in $\Lambda_{i}^M$.
 
 Finally, the \textit{sample-level} calibration for a normalized support sample $\bar{x}_i$ is defined as:
 \begin{equation}
  \label{eq:sample-calib}
  \bar{x}_i^s = normalize((1- \alpha) \bar{x}_i  + \alpha{\bar{s}_i}),
 \end{equation}
 
 where $\alpha \in [0,1]$ is a parameter to linearly combine the normalized support feature $\bar{x}_i$ and normalized base-prototypes-driven calibration $\bar{s}_i = norm (s_i)$.
 As shown in Figure \ref{framework}, $\bar{x}_i$ and $\bar{s}_i$ form a line in the normalized feature space and $\bar{x}_i^s$ is the normalization of a in-between point on this line.
 In general, the sample-level calibration can rectify each support sample based on its own top $M$ most similar base classes.
 
 \subsubsection{Task-Level Calibration}\label{Task}
 By performing the sample-level calibration, the bias induced by the few-shot support samples can be reduced to a certain degree.
 However, when the sampling bias is too large, e.g., the support sample is lying near the boundary of a class, the set of similar base classes $\Lambda_{i}^M$ obtained by Eq. \ref{eq:k-nearest} may also be biased.
 To alleviate such bias, we propose the \textit{task-level} calibration which utilizes the base prototypes related to all support samples when calibrating each individual support feature.
 Concretely, for a support set $\mathbf{S} = \{ (x_i, y_i)\}_{i=1}^{N \times K}$ w.r.t a task $\mathcal{T}$, we collect the top $M$ similar base prototypes for each support sample and form a union of related base prototypes for $\mathcal{T}$:
 \begin{equation}
  \label{eq:10}
  \Lambda_{\mathcal{T}} = \bigcup\limits_ {i=1 }^{N \times K}\Lambda_{i}^M.
 \end{equation}
 
 Then, for a transformed support sample $\tilde{x}_i$ obtained by Eq. \ref{eq:transformation}, the calibration using all of the task-related base prototypes is computed by:
 \begin{equation}
  \label{eq:task-pb}
  t_i = \tilde{x}_i +\sum_{j \in \Lambda_{\mathcal{T}} } w_{ij} p_j^b,
 \end{equation}
 where $w_{ij}$ is calculated in the similar way as Eq. \ref{eq:weight}, but the similarities are computed using the prototypes from $\Lambda_{\mathcal{T}}$ instead of $\Lambda_{i}^M$.
 By involving more prototypes to calibrate the support samples, the bias caused by only using nearby prototypes for a near-boundary support sample can be reduced.
 
 Then, we define the task-level calibration for a normalized support sample $\bar{x}_i$ as:
 \begin{equation}
  \label{eq:task-calib}
  \bar{x}_i^t = normalize((1- \beta) \bar{x}_i  + \beta \bar{t}_i),
 \end{equation}
 where $\bar{t}_i$ is the normalization of $t_i$.
 Similar to the sample-level calibration, $\bar{x}_i$ and $\bar{t}_i$ also form a line in the normalized feature space, while the calibration for each support sample is based on the union of all related base prototypes $\Lambda_{\mathcal{T}}$.

 \subsubsection{Unified Model}\label{Unified}
 The sample-level and task-level calibration utilize different levels of information from the base classes to rectify the support samples in a discrete manner.
 To further attain the merits of both calibration schemes, we propose a unified model which linearly combines the sample-level and task-level calibration:
 \begin{equation}
  \label{eq:final-calib}
  x_i^c=  \bar{x}_i^u = normalize((1 - \alpha - \beta) \bar{x}_i + \alpha \bar{s}_i + \beta \bar{t}_i).
 \end{equation}
 Here, $\bar{x}_i^u$ which is also denoted as $x_i^c$, is the final calibration for a normalized support sample $\bar{x}_i$  .
 Geometrically, $x_i^c$ can be understood as the normalization of an interpolated feature point $x_i^u$ locating in the triangle formulated by the three vertices $\bar{x}_i$, $\bar{s}_i$ and $\bar{t}_i$, while $1-\alpha - \beta$, $\alpha$ and $\beta$ are the barycentric coordinates of $x_i^u$.
 Different $\alpha$ and $\beta$ values can lead to different calibration effects.
 When $\beta=0$, the unified model degenerates to the sample-level calibration, while when $\alpha=0$, the model becomes to the task-level calibration.
 We quantitatively evaluate the effects of different $\alpha$ and $\beta$ values in Section \ref{sec:ablation}.
 
 \subsection{Nearest Prototype Classifier}\label{NPC}
 With the calibrated support set $\mathbf{S}^c = \{(x_i^c, y_i)\}_{i=1}^{N \times K}$, we compute the prototypes $\{p_n\}_{n=1}^{N}$ for the novel classes and perform cosine similarity based nearest classification for a query feature $q$.
 To simplify the notation, we further represent $\mathbf{S}^c = \{\mathbf{S}_n^c\}_{n=1}^N$, while $\mathbf{S}_n^c = \{(x_k^c, y_k=n)\}_{k=1}^K$ is the support set for a novel class $CLS_n$.
 
 For the 1-shot case, each calibrated support sample becomes one prototype and the class of the query feature is predicted by the nearest prototype classifier:
 \begin{equation}
  \label{eq:nn}
  y^* = \max_{p_n} cos(\bar{q}, p_n),
 \end{equation}
 where $p_n = x_n^c$ is the calibrated prototype for novel class $CLS_n$ and $\bar{q}$ is the normalization of query q.
 
 For the multi-shot case, one way to obtain the prototype for a novel class is simply to compute the average of all support features for the given class as in Prototypical Networks \cite{prototypical}.
 However, merely using the unweighted average of the support features as prototype does not consider the importance of the support samples w.r.t the query.
 Therefore, we adopt the idea of \textit{attentive prototype} which is proposed in recent works \cite{wu2020,wang2022} for query-guided prototype computation.
 In our implementation, we define the attention-weighted prototype as: 
 \begin{align}
  p_n^q &=  \sum_{x_k^c \in \mathbf{S}_n^c} a_k x_k^c,  \label{eq:query-proto} \\
  where \enspace a_k &= \frac{e^{<q, x_k^c>}}{\sum_{x_m^c \in \mathbf{S}_n^c}e^{<q, x_m^c>}}.
 \end{align}
 Here, $x_k^c$ and $x_m^c$ are the calibrated support samples belonging to the $CLS_n$'s support set
 $\mathbf{S}_n^c$ and $a_k$ is the attention weight computed by applying Softmax to the similarities between query $q$ and these calibrated support samples;
 $p_n^q$ is the $CLS_n$'s prototype guided by query $q$.
 Similar to Eq. \ref{eq:nn}, the prediction for a query $q$ is obtained by finding the novel class with the nearest prototype $p_n^q$.
 
 \section{Experiments}
 In this section, we perform quantitative comparisons between our P3DC-Shot and state-of-the-art few-shot classification methods on three representative datasets.
 We also conduct ablation studies on evaluating different hyperparameters and design choices for our methods.
 Our code is available at: https://github.com/breakaway7/P3DC-Shot.
 
 \begin{table*}[t]
  \centering
  \small
  \caption{ Quantitative comparison on the test set of miniImageNet, tieredImageNet and CUB. The 5-way 1-shot and 5-way 5-shot classification accuracy (\%) with 95\% confidence intervals are measured. Best results are highlighted in bold and second best are in italic. The last line shows the $\alpha$ and $\beta$ selected based on the valiation set for each dataset.
   * 8 and 20 are the number of ensembles in DeepVoro and DeepVoro++.
   $\dagger$ The results of \cite{xu2022alleviating} on tieredImageNet are obtained using its released code.
  }
  \centering
  \setlength{\tabcolsep}{4.5pt}
  \label{tab:mini_cub_result}
  \begin{tabular}{l|cc|cc|cc}
   \hline
   \multirow{2}{*}{Methods} &\multicolumn{2}{c}{\textit{miniImageNet}} 
   & \multicolumn{2}{c}{\textit{tieredImageNet}}
   & \multicolumn{2}{c}{\textit{CUB}} \\
   &{5-way 1-shot} & {5-way 5-shot}  &{5-way 1-shot} & {5-way 5-shot}  &{5-way 1-shot} & {5-way 5-shot} \\
   \hline
   
   \multicolumn{4}{l}{\textbf{\textit{Meta-learning (metric-learning)}}}\\
   MatchingNet \cite{matching} (2016)   &$64.03 \pm 0.20$ &$76.32\pm0.16$
   &$68.50\pm0.92$ &$80.60\pm0.71$
   &$73.49\pm0.89$ &$84.45\pm0.58$ 
   \\
   
   ProtoNet \cite{prototypical} (2017)   &$54.16 \pm0.82$ &$73.68\pm0.65$ 
   &$65.65\pm0.92$ &$83.40\pm0.65$ 
   &$72.99\pm0.88$ &$86.64\pm0.51$ \\
   
   RelationNet \cite{relation} (2018)  &$52.19 \pm0.83$ &$70.20\pm0.66$  
   &$54.48\pm0.93$ &$71.32\pm0.78$
   &$68.65\pm0.91$ &$81.12\pm0.63$\\

   \hline
   \multicolumn{4}{l}{\textbf{\textit{Meta-learning (optimization)}}}\\
   MAML \cite{MAML} (2017)   &$48.70 \pm1.84$ &$63.10\pm0.92$ 
   &$51.67\pm1.81$ &$70.30\pm0.08$ 
   &$50.45\pm0.97$ &$59.60\pm0.84$ 
   \\
   
   LEO \cite{LEO} (2019)   &$61.76 \pm0.08$ &$77.59\pm0.12$ 
   &$66.33\pm0.15$ &$81.44\pm0.09$
   &$68.22\pm0.22$ &$78.27\pm0.16$ 
   \\
   
   DCO \cite{DCO} (2019)   &$62.64 \pm0.61$ &$78.63\pm0.46$ 
   &$65.99 \pm0.72$ &$81.56\pm0.53$ 
   &- &-  
   \\
   
   \hline
   \multicolumn{4}{l}{\textbf{\textit{Transfer learning}}}\\
   Baseline++ \cite{closer} (2019)
   &$57.53\pm0.10$ &$72.99\pm0.43$ 
   & $60.98\pm0.21$ &$75.93\pm0.17$
   & $70.40\pm0.81$ &$82.92\pm0.78$
   \\

   Negative-Cosine \cite{negative} (2020)
   &$62.33\pm0.82$ &$80.94\pm0.59$ 
   & - &-
   & $72.66\pm0.85$ &$89.40\pm0.43$  \\
   
   $\mathrm{S2M2_{R}}$ \cite{s2m2} (2020) 
   &$64.65 \pm0.45$ &$83.20 \pm0.30$ 
   & $68.12 \pm0.52$ &$86.71 \pm0.34$
   & $80.14 \pm0.45$ &$90.99\pm0.23$ 
   \\
   
   \hline
   
   \multicolumn{4}{l}{\textbf{\textit{Nearest neighbor}}}\\
   SimpleShot \cite{simpleshot} (2019)
   &$64.29\pm0.20 $ &$81.50\pm0.14$
   &$71.32\pm0.22 $ &$86.66\pm0.15$
   & - &-  \\
   
   DeepVoro$(8)^{*}$ \cite{CIVD} (2022)  
   &$66.45\pm0.44 $ &$\textbf{84.55}\pm \textbf{0.29}$ 
   & $74.02 \pm0.49$ &$\textbf{88.90}\pm \textbf{0.29}$ 
   & $80.98\pm0.44$ &$\textit{91.47}\pm\textit{0.22}$  
   \\
   
   DeepVoro++$(20)^{*}$ \cite{CIVD} (2022)
   &${68.38}\pm{0.46} $ & $83.27\pm0.31$ 
   & $\textit{74.48} \pm\textit{0.50}$ &- 
   & $80.70\pm0.45$ &-
   \\
   
   \hline
   
   \multicolumn{4}{l}{\textbf{\textit{Data calibration}}}\\

   RestoreNet \cite{one-shot} (2020)
   &$59.28 \pm0.20$ &-  
   &- &-
   &$74.32 \pm 0.91$  &-    \\
   
   DC \cite{DC} (2021)  &$67.79\pm0.45 $ &$83.69\pm0.31$ 
   & ${74.24}\pm{0.50}$ &$88.38\pm0.31$
   & $79.93\pm0.46$ &$90.77\pm0.24$  
   \\
   
   MCL-Katz+PSM \cite{guo2022learning} (2022)
   &$67.03 $ &$84.03$ 
   & $69.90$ &$85.08$
   & $\textbf{85.89}$ &$\textbf{93.08}$  
   \\
   
   S2M2+TCPR$^\dagger$ \cite{xu2022alleviating} (2022)
   &${68.05}\pm{0.41}$ &$\textit{84.51}\pm\textit{0.27}$
   &$72.67\pm0.48$ &$87.96\pm0.31$
   &- &-\\
   
   \hline
   
   P3DC-Shot ($\alpha=0$, $\beta=0$)
   &${65.93}\pm{0.45}$ &$84.06\pm0.30$
   &${73.56}\pm{0.49} $ &${88.50}\pm{0.32}$ 
   & ${81.61}\pm{0.43}$ &$91.36\pm0.22$
   \\
   
   P3DC-Shot ($\alpha=1$, $\beta=0$)
   &${68.41}\pm{0.44}$ &$83.06\pm0.32$
   &${74.84}\pm{0.49} $ &${88.01}\pm{0.33}$ 
   & ${81.51}\pm{0.44}$ &$90.83\pm0.24$
   \\
   
   P3DC-Shot ($\alpha=0$, $\beta=1$)
   &$\textit{68.67}\pm\textit{0.44}$ &$83.64\pm0.31$
   &$\textbf{75.20}\pm\textbf{0.48} $ &${88.29}\pm{0.33}$ 
   & ${81.58}\pm{0.44}$ &$91.02\pm0.23$
   \\
   
   P3DC-Shot ($\alpha=\frac{1}{3}$, $\beta=\frac{1}{3}$)
   &${68.33}\pm{0.44}$ &$84.19\pm0.30$
   &$\textit{74.91}\pm\textit{0.49} $ &${88.54}\pm{0.32}$ 
   & ${81.75}\pm{0.43}$ &$91.21\pm0.23$
   \\
   
   P3DC-Shot (selected $\alpha$, $\beta$)
   &$\textbf{68.68}\pm\textbf{0.44}$ &$84.37\pm0.30$
   &$\textbf{75.20}\pm\textbf{0.48} $ &$\textit{88.67}\pm\textit{0.32}$ 
   & $\textit{81.86}\pm\textit{0.43}$ &$91.36\pm0.23$
   \\
   
   &$(0.0, 0.9)$ 
   &$(0.0, 0.4)$
   &$(0.0, 1.0)$ 
   &$(0.0, 0.3)$
   &$(0.2, 0.4)$
   &$(0.0, 0.4)$
   \\

   \hline
  \end{tabular}
  \label{comp-sota}
 \end{table*}
 
 \subsection{Datasets}
 We evaluate our prior-driven data calibration strategies on three popular datasets for benchmarking few shot classificaiton: miniImageNet \cite{2015ImageNet}, tieredImageNet \cite{M2018Meta} and CUB \cite{CUB}. 
 miniImageNet and tieredImageNet contain a broad range of classes including various animals and objects, while CUB is a more fine-grained dataset that focuses on various species of birds. 
 
 Specifically, the miniImageNet \cite{2015ImageNet} is derived from the ILSVRC-2012 \cite{imagenet} and it contains a subset of 100 classes, each of which consisting of 600 images. 
 We follow the split used in \cite{optimization} and obtain 64 base, 16 validation and 20 novel classes for miniImageNet.
 Comaring to miniImageNet, the tieredImageNet \cite{M2018Meta} is a larger subset of \cite{imagenet} which contains 608 classes and therefore more challenging. 
 We follow \cite{M2018Meta} and split the tieredImageNet into 351, 97, and 160 classes for base, validation, and novel classes, respectively.
 For CUB \cite{CUB}, it is the short name for Caltech-UCSD Birds 200 dataset, which contains a total of 11,788 images covering 200 categories of different bird species. 
 We split the CUB dataset into 100 base, 50 validation and 50 novel classes following \cite{closer}.
 Note that the set formed by the base classes can also be regarded as the train set and the novel classes correspond to the test set.

 \subsection{Implementation Details}
 For each image in the dataset, we represent it as a 640-dimensional feature vector which is extracted using the WideResNet \cite{wide} pretrained by the S2M2 \cite{s2m2} work.
 Our calibration pipeline can efficiently proceed in four steps:
 1) find the $M=5$ nearby base prototypes for each support sample $x_i$;
 2) compute the endpoint of the sample-level calibration for $x_i$, i.e., $s_i$;
 3) collect all nearby base prototypes for all support samples in the task and compute the endpoint of the task-level calibration for $x_i$, i.e., $t_i$;
 4) combine the sample-level and task-level calibration and obtain the final calibrated support sample $x_i^c$.
 The parameter $\alpha$ and $\beta$ for weighting the sample-level and task-level calibration are selected based on the best results obtained on the validation set for each dataset.
 All experiments are conducted on a PC with a 2.70GHz CPU and 16G memory.
 No GPU is needed during the calibration.
 On average, for a 5-way 5-shot task, it takes 0.027 seconds to calibrate the support samples
 and 0.002 seconds for performing the nearest prototype classification.
 
 \subsection{Comparison and Evaluation}
 To evaluate the performance of our P3DC-Shot, we first conduct quantitative comparisons with some representative and state-of-the-art few-short classification methods.
 Then, we compare with different data transformation or calibration schemes and provide qualitative visualization for showing the difference of our calibration results w.r.t existing works.
 In addition, we evaluate the generalizability of our method by performing classification tasks with different difficulties. 
 
 \textbf{Quantitative comparisons.}
 As there are numerous efforts have been paid to the few-shot classification, we mainly compare our P3DC-Shot with representative and SOTA works which cover different types of few-shot learning schemes.
 The compared methods include the metric-learning based meta-learning \cite{matching,prototypical,relation}, optimization-based meta-learning \cite{MAML,LEO,DCO}, transfer learning \cite{closer,negative,s2m2}, nearest neighbor \cite{simpleshot,CIVD} and calibration \cite{one-shot,DC,guo2022learning,xu2022alleviating} based methods.
 For certain methods such as \cite{s2m2,ma2022few}, we only compare with their basic versions and do not consider their model trained with data augmentation.
 Note that as not every method has conducted experiments on all three datasets, we mainly compare with their reported results.
 One exception is for \cite{xu2022alleviating}, we compare with its results generated using its released code.
 
 For our method, we report the results of our model with different hyperparameters $\alpha$ and $\beta$.
 In particular, we consider the case when $\alpha$ and $\beta$ are both zero, which makes our method a simple NN-based method with no data calibration and only shows the effect for using the query-guided prototype computation (Eq. \ref{eq:query-proto}).
 We also compare with the results of $\alpha$ or $\beta$ is 1, or both of them are equal to $\frac{1}{3}$, which correspond to the cases that the endpoint of the sample-level or task-level calibration or the barycenter of the calibration triangle (Figure \ref{framework}).
 In the end, we provide our best results with the $\alpha$ or $\beta$ selected based on the validation set. 
 
 \begin{figure*}[t]
  \centering
  \includegraphics[width=0.92\linewidth]{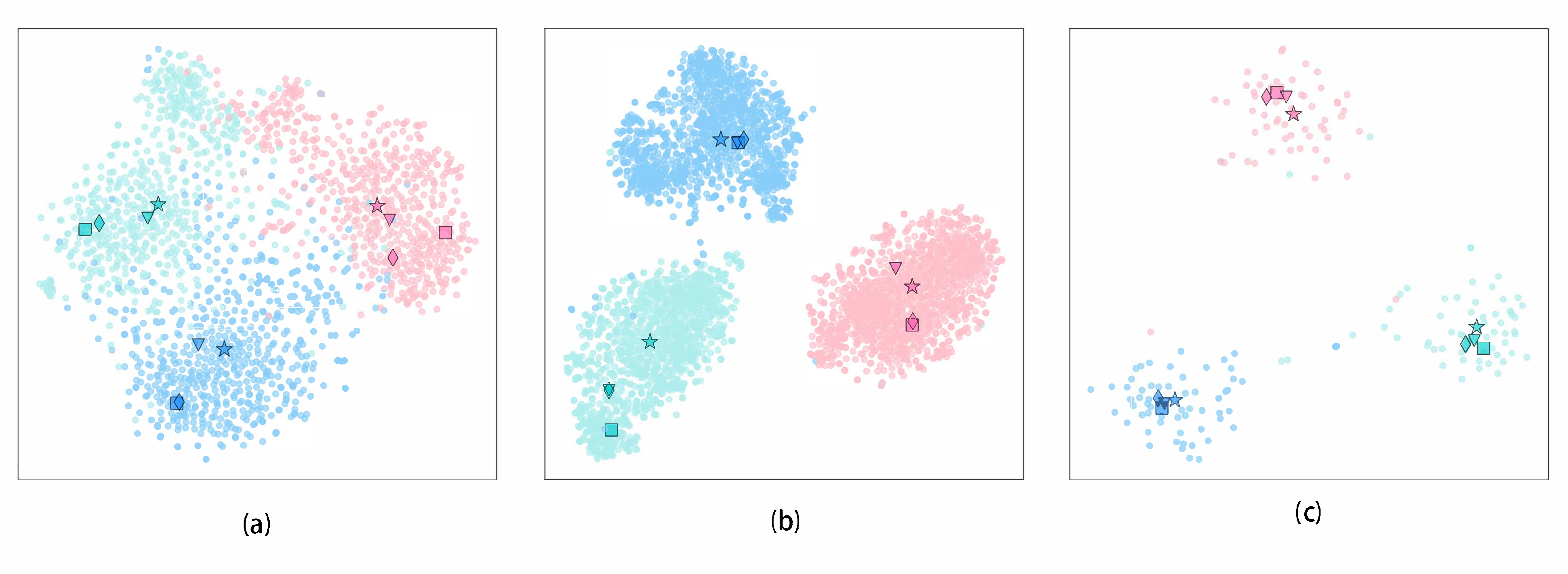}  
  \caption{ T-SNE visualization of the calibration on example support samples from the test set of miniImageNet (a), tieredImageNet (b), and CUB (c). 
   The colored dots are data from the same underlying classes as the selected sample and the star is the center of each class. 
   Given a support sample (represented in square), the upside down triangle is our calibration result and the lozenge is the calibration result of DC \cite{DC}.
  }
  \label{tsne}
 \end{figure*}
 
 For each dataset, we evaluate on the 5-way 1-shot and 5-way 5-shot classification setting.
 For each setting, 2,000 testing tasks, each of which contains $5\times K$ ($K=1 \enspace or \enspace 5$) samples for the support set and $5 \times 15$ samples for the query set, are randomly generated from the test split of the corresponding dataset.
 Table \ref{comp-sota} shows the quantitative comparison results on three datasets.
 It can be seen that our best results outperform most methods in the 5-way 1-shot setting and are comparable to the SOTA methods \cite{ma2022few, DC} for the 5-way 5-shot setting.
 Note that although \cite{guo2022learning} achieves best results on the CUB dataset, it is inferior on miniImageNet and tieredImageNet.
 Moreover, since \cite{guo2022learning} follows a metric-based few-shot learning pipeline, it still requires to train the feature extractor and the metric module for each dataset.
 For \cite{ma2022few}, it performs generally well on all three datasets, but as an ensemble-based method, its computation time is much longer than our method, especially when the ensemble number is large.
 In contrast, our method does not require any training and only needs to perform an efficient calibration step for each testing task.
 
 Also, from results of our method with different $\alpha$ and $\beta$ values in Table 1, it can be found when $\alpha$ and $\beta$ is zero, the query-guided prototype computation can lead to better performance than the simple NN-based SimpleShot \cite{simpleshot}.
 When either the sample-level or task-level calibration is applied, i.e., $\alpha$ or $\beta$ is not zero, the results are better than the non-calibrated version, showing the calibration can indeed reduce the bias for the support samples.
 Meanwhile, which calibration type is more suitable is depending on the underlying data distribution of the dataset.
 By selecting the $\alpha$ and $\beta$ based on the validation set of each dataset, the results are further improved.
 In the ablation study, we perform more experiments and analysis of different $\alpha$ and $\beta$ values.
 
 \begin{table}[h]
  \centering
  \caption{Comparison with different data transformation or calibration schemes. Accuracy (\%) for 5-way 1-shot task on the test set of miniImageNet are measured.} 
  \label{tab:diff-calib}
  \begin{tabular}{lc|cc}
   \hline
   \multirow{2}{*}{Model}& 
   &miniImageNet &CUB\\
   \cline{3-4}
   &&5-way 1-shot & 5-way 1-shot  \\
   \hline
   NN & &${47.50} $ &${76.40} $\\
   L2N+NN& &${65.93} $&${81.61} $\\
   
   CL2N+NN& &${65.96} $&${81.54} $\\
   
   DC+L2N+NN& &${66.23} $&${79.49} $\\
   
   \hline
   {P3DC-Shot} &&$\textbf{68.68} $&$\textbf{81.86}$  \\
   (selected $\alpha$, $\beta$) && (0.0,0.9) & (0.2,0.4)  \\

   \hline
  \end{tabular}
 \end{table}

 \begin{table*}[t]
  \caption{Generalizability test on different N in N-way 1-shot tasks. Accuracy (\%) on the test set of miniImageNet are measured. For our P3DC-Shot, the same $\alpha=0$ and $\beta=0.9$ selected based on the validation set for the 5-way 1-shot case are used for all experiments.}
  \begin{center}
   \scalebox{0.95}{
    \begin{tabular}{l|ccccccc}
     \hline
     Models                            & 5-way & 7-way & 9-way & 11-way & 13-way & 15-way & 20-way \\
     \hline
     
     RestroreNet \cite{one-shot} & $59.56$ & $50.55$ & $44.54$ & $39.98 $  & $36.34$  & $33.52$  & $28.48$\\
     \hline
     L2N+NN  &$65.93$  &$57.86$  &$52.45$ &$48.25$ 
     &$44.80$ &$42.12$  &$37.06$ \\   \hline
     CL2N+NN  &$65.96$  &$57.69$  &$52.23$ &$47.93$ 
     &$44.36$ &$41.85$  &$36.65$ \\   
     \hline
     \textbf{P3DC-Shot}  &$68.68$ &$60.58$  &$55.03$ &$50.75$ 
     &$47.21$ &$44.43$  &$39.33$ \\   \hline 
     
     \hline
    \end{tabular}
   }
  \end{center}
  \label{nway}
 \end{table*}

 \textbf{Comparison with different data transformation or calibration schemes.}
 To further verify the effectiveness of our prior-driven data calibration, we compare with several NN-based baseline methods which perform different data transformation or calibration schemes and the results are shown in Table \ref{tab:diff-calib}.
 In this experiment, all methods are based on the pretrained WideResNet features.
 Also, only the 5-way 1-shot classification accuracy is measured so that the comparison is focused on feature transformation instead of the prototype computation schemes.
 The first baseline is NN, which is a naive inner product based nearest neighbor classifier.
 Then, L2N and CL2N represent L2 normalization and centered L2 normalization which have been shown as effective in SimpleShot \cite{simpleshot}.
 In addition, another baseline that follows the data calibration scheme in DC \cite{DC} is compared.
 Comparing to the original DC, this baseline directly takes the calibrated and then normalized features and employs NN for classification instead of training new classifiers using the sampled data.
 From Table \ref{tab:diff-calib}, it can be observed the data normalization or calibration can significantly improve the NN-based classification.
 In addition, our data calibration achieves the best results comparing to other baselines.
 The main reason is the L2N and CL2N only perform transformation rather than calibration using the base priors, while the modified DC does not consider the attentive similarity between the support samples and the base classes when performing the calibration.

 \textbf{Visualization of the calibration.}
 To qualitatively verify the effectiveness of our calibration, we show the T-SNE \cite{Maaten2008VisualizingDU} visualization of the calibration results for some example support samples in Figure \ref{tsne}.
 The results of calibrating the same sample using DC \cite{DC} are also compared.
 It can be seen from Figure \ref{tsne} that our calibration can more effectively transform the support samples closer to the center of the underlying classes.
 For DC, the calibration may be minor or even be far away from the center.
 The reason is still due to it treats the nearby base classes with the same weights.
 In contrast, our calibration pays more attention to the similar base classes when determining the weights for combining the base prototypes (Eq. \ref{eq:sample-pb} and \ref{eq:task-pb}).
 
 \textbf{Generalizability test on different N in N-way classification.} 
 Following \cite{one-shot}, we conduct a series of N-way 1-shot experiments on miniImageNet to test the generalizability of the proposed calibration for different classification tasks.
 Table \ref{nway} shows the results of the baseline methods \cite{one-shot}, L2N and CL2N and ours.
 Note that with the N increases, there are more data samples in a test task and the classification becomes more difficult.
 It can be observed that our P3DC-Shot achieves consistent best results comparing to the baseline methods, verifying our method is generalizable to classification tasks with different difficulties.

  \begin{figure*}[!t]
  \centering
  \includegraphics[width=0.92\linewidth]{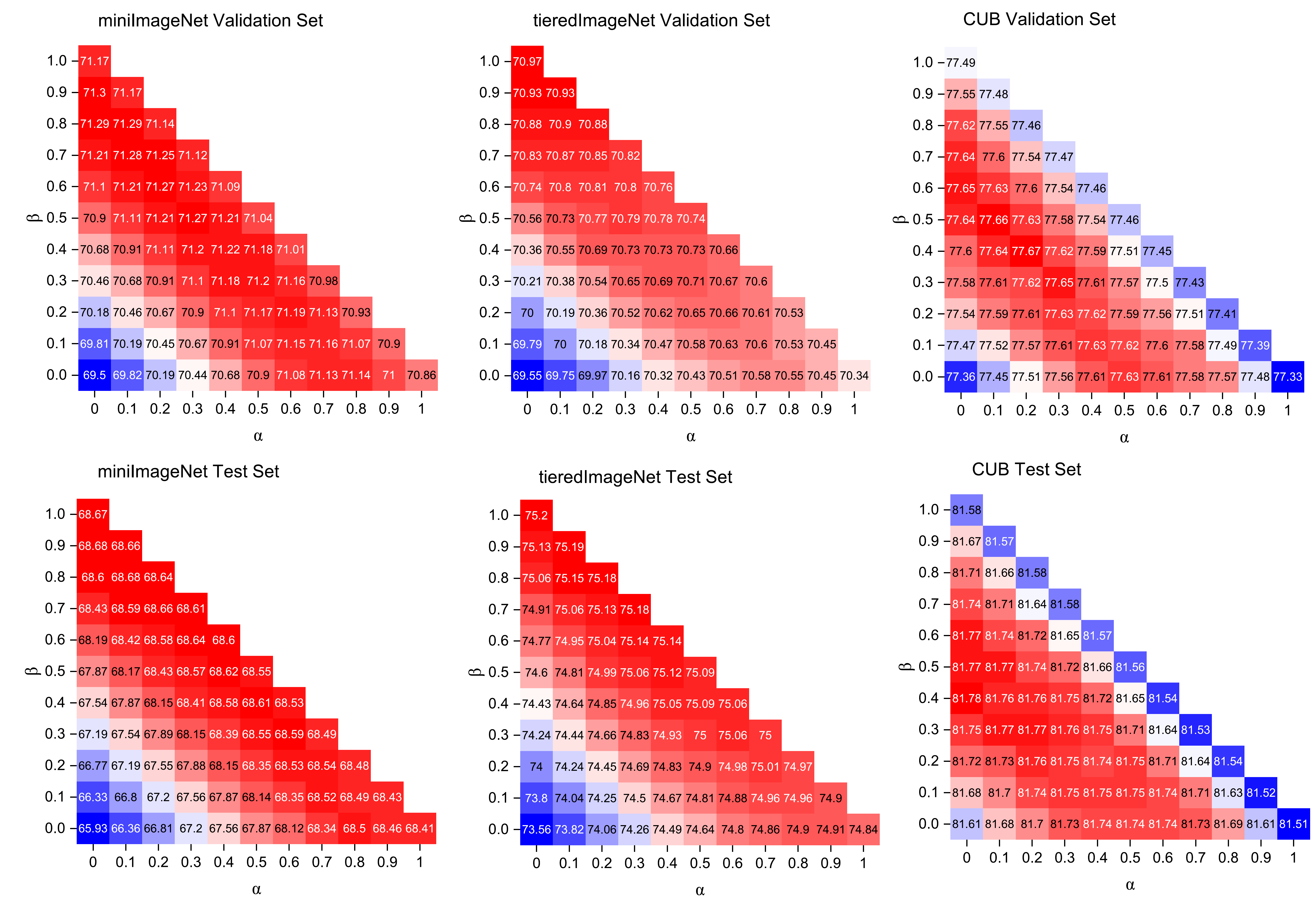}
  \caption{The effect of different $\alpha$ and $\beta$ on the validation (top) and test (bottom) set of different datasets. Accuracy (\%) for 5-way 1-shot task on miniImageNet, tieredImageNet and CUB are measured.
   The warmer color corresponds to higher accuracy.
  }
  \label{hyperparameter}
 \end{figure*}
 
 \subsection{Ablation Study}
 \label{sec:ablation}
 
 In this section, we perform ablation studies to verify the effectiveness of different modules and design choices of our method.
 First, we conduct experiments on different hyperparameter $\alpha$ and $\beta$ to see how the sample-level and task-level calibration can affect the final results.
 Then, we perform the study on the effectiveness of using the query-guided attentive prototypes in the NN classification step.
 
 \textbf{Effect on different hyperparameter $\alpha$, $\beta$.}
 Different $\alpha$ and $\beta$ values correspond to different degrees of sample-level and task-level calibration applied to the input data.
 Geometrically, $\alpha$, $\beta$ and $1-\alpha-\beta$ can also be understood as the coordinates of the calibration result w.r.t to the triangle formed by the three points $\bar{x}_i, s_i, t_i$.
 To quantitatively reveal how these two hyperparameters can affect the results, we enumerate different $\alpha$ and $\beta$ values on both the validation and test sets of different datasets.
 From the results in Figure \ref{hyperparameter}, it can be found the accuracy near the origin of the figures are smaller, which means performing calibration can improve upon using the original features for classification, i.e., $\alpha$ and $\beta$ is zero.
 Also, different datasets prefer different $\alpha$ and $\beta$ combinations for achieving higher performance.
 For example, miniImageNet shows better results when $\alpha + \beta$ is around 0.9 and CUB prefers a relatively smaller calibration, i.e., $\alpha + \beta$ is around 0.6.
 For tieredImageNet, better results are obtained around the topper left of the figure, showing the task-level calibration is more helpful than the sample-level.
 Overall, the trend on the test set is consistent with the validation set.
 From above experiments, it shows the sample-level and task-level calibration are consistently effective, while how to selecting the good $\alpha$ and $\beta$ values are dataset dependent.
 Therefore, for our best results, we use the $\alpha$ and $\beta$ selected based on the validation set and report their performance on the test set.

 \textbf{Effect on using attentive prototypes in NN classification.}
 To improve the conventional prototype based NN classificaiton, we propose to compute the query-guided attentive prototypes to represent the support class.
 To verify the effectiveness of this scheme, we perform ablation study for 5-way 5-shot tasks on different tasks using different prototype computation schemes.
 Specifically, we take the calibrated support features and compute the prototypes for the support classes by performing the conventional average operation or our query-guided attentive averaging (Eq. \ref{eq:query-proto}).
 The results in Table \ref{tab:query-proto} show that the attentive prototypes can lead to better performance.
 Hence, we adopt the attentive prototypes in our NN-based classification.
 
 \begin{table}[t]
  \centering
  \caption{Ablation study on using the query-guided attentive prototypes in NN classification.
   Accuray (\%) on the test set of miniImageNet, tieredImageNet and CUB are measured.
  } 
  \label{tab:query-proto}
            \scalebox{0.9}{
            \setlength{\tabcolsep}{2pt}
   \begin{tabular}{lc|ccc}
    \hline
    \multirow{2}{*}{Model}& 
    &miniImageNet &tieredImageNet &CUB\\
    &&5-way 5-shot & 5-way 5-shot   & 5-way 5-shot \\
    \hline
    Average & &${84.11} $ &${88.54}$ & ${91.27} $\\
    Attentive & &${84.37} $ &${88.67}$ &${91.36}$\\
    
    \hline
   \end{tabular}
   }
 \end{table}

 \section{Conclusion}
 In this paper, we propose a simple yet effective framework, named P3DC-Shot, for few-shot classification.
 Without any retraining and expensive computation, our prior-driven discrete data calibration 
 method can efficiently calibrate the support samples based on prior-information from the base classes to obtain the less-biased support data for NN-based classification.
 Extensive experiments show that our method can outperform or at least comparable to SOTA methods which need additional learning steps or more computation.
 One limitation of our method is we rely on the whole validation set to select the good hyperparameters $\alpha$ and $\beta$ to determine which degree of the sample-level and task-level calibration is more suitable for the given dataset.
 Investigating a more general scheme to combine the sample-level and task-level calibration is an interesting future work.
 Moreover, when exploring the combination schemes, we only focus on exploring the inner area of the calibration triangle.
 It is worthy to extend the parameter search to a larger area, i.e., by extrapolation of the calibration triangle, to find whether better results can be obtained.

 
 
 
 \bibliographystyle{elsarticle-num-names}

 \bibliography{cas-refs}
 
  
  
  
\end{document}